\documentclass[pdflatex,sn-mathphys-num]{sn-jnl}
\usepackage{graphicx}%
\usepackage{multirow}%
\usepackage{amsmath,amssymb,amsfonts}%
\usepackage{amsthm}%
\usepackage{xcolor}%
\usepackage{textcomp}%
\usepackage{manyfoot}%
\usepackage{booktabs}%
\usepackage{algorithm}%
\usepackage{algorithmicx}%
\usepackage{algpseudocode}%
\usepackage{listings}%
\usepackage{subcaption}
\usepackage{makecell}
\usepackage{rotating}

\begin{document}

\title{\textbf{Randomness and signal propagation in physics-informed neural networks (PINNs):  A neural PDE perspective}}
\author[1]{\fnm{Jean-Michel} \sur{Tucny}}\email{jeanmichel.tucny@uniroma3.it}

\author*[2]{\fnm{Abhisek} \sur{Ganguly}}\email{abhisek@jncasr.ac.in}

\author[2]{\fnm{Santosh} \sur{Ansumali}}\email{ansumali@jncasr.ac.in}
\author[3]{\fnm{Sauro} \sur{Succi}}\email{sauro.succi@iit.it}

\affil[1]{\orgdiv{Department of Civil, Computer Science and Aeronautical Technologies Engineering}, 
          \orgname{Università degli Studi Roma Tre}, 
          \orgaddress{\street{via Vito Volterra 62}, \city{Rome}, \postcode{00146}, \state{RM}, \country{Italy}}}      
\affil[2]{\orgdiv{Engineering Mechanics Unit}, 
          \orgname{Jawaharlal Nehru Centre for Advanced Scientific Research}, 
          \orgaddress{\street{Jakkur}, \city{Bangalore}, \postcode{560064}, \state{Karnataka}, \country{India}}}
\affil[3]{\orgdiv{Center for Life Nano-\& Neuro-Science}, 
          \orgname{Italian Institute of Technology (IIT)}, 
          \orgaddress{\street{viale Regina Elena 295}, \city{Rome}, \postcode{00161}, \state{RM}, \country{Italy}}}

\abstract{
Physics-informed neural networks (PINNs) often exhibit weight matrices that appear statistically random after training, yet their implications for signal propagation and stability remain unsatisfactorily understood, let alone the interpretability.
In this work, we analyze the spectral and statistical properties of trained PINN weights using viscous and inviscid variants of the one-dimensional Burgers’ equation, and show that the learned weights reside in a high-entropy regime consistent with predictions from random matrix theory.
To investigate the dynamical consequences of such weight structures, we study the evolution of signal features inside a network through the lens of neural partial differential equations (neural PDEs).  We show that random and structured weight matrices can be associated with specific discretizations of neural PDEs, and that the numerical stability of these discretizations governs the stability of signal propagation through the network. 
In particular, explicit unstable schemes lead to degraded signal evolution, whereas stable implicit and higher-order schemes yield well-behaved dynamics for the same underlying neural PDE.
Our results offer an explicit example of how numerical stability and network architecture shape signal propagation in deep networks, in relation to random matrix and neural PDE descriptions in PINNs.
}
\maketitle

\section{Introduction}

The rapid adoption of neural computing and machine learning has transformed scientific computing, enabling major breakthroughs in fields ranging from protein folding to weather prediction~\cite{jumper2021highly,bi2023accurate}. 
Despite these successes, most modern models operate as high-dimensional black boxes, with their decision-making processes encoded in millions of learned parameters, limiting physical interpretability and mechanistic insight.
This opacity is closely tied to current training paradigms, which rely heavily on large, weakly structured weight matrices and brute-force optimization in high-entropy regimes~\cite{succi2026neuralchains}. 
Although architectural choices may sometimes encode spatial or temporal structure~\cite{lecun1998gradient}, most approaches lack explicit connections between learned weights and underlying physical laws. Such approaches are computationally expensive and contribute to rapidly growing energy demands in large-scale machine learning systems~\cite{masanet2020recalibrating,jones2018energy}, raising concerns about long-term sustainability. These challenges motivate a deeper theoretical understanding of neural network weight matrices and their role in governing signal propagation, stability, and convergence. In particular, developing principled frameworks that connect learned weights to underlying physical structures, dynamical mechanisms, and conservation laws may enable more efficient and interpretable scientific machine learning models, driven by insight.

Random matrix theory (RMT) provides a useful statistical framework for analyzing the spectral properties of neural network weight~\cite{MengYao2023,MartinMahoney2018}. 
Previous studies have shown that both randomly initialized and trained networks often exhibit universal spectral behavior governed by Marchenko-Pastur and circular laws, with task-specific information localized in spectral outliers~\cite{louart2018random,pennington2017nonlinear,thamm2022random}. 
These results establish randomness and universality as important features of weight statistics, but do not fully explain the resulting network dynamics.

Recent work has shown that signal propagation in deep networks can be interpreted through neural partial differential equations (neural PDEs), which describe the layer-wise evolution of activations as discretizations of underlying continuous dynamical systems~\cite{NPDE,succi2026neuralchains}. Closely related continuous-depth formulations, such as neural ODEs~\cite{chen2019neural}, provide an alternative perspective in which depth is modeled explicitly as a time variable. Within this framework, network behavior is governed not only by weight statistics, but also by the numerical properties of the implicit time-stepping schemes implemented by the architecture.

From this dynamical perspective, stability and convergence of neural networks are closely related to classical concepts from numerical analysis~\cite{alt2022connections}, including explicit and implicit integration schemes, fixed-point convergence, and equilibrium dynamics~\cite{hu2022spectrum,bai2019deep,rubanova2019latent}. 
These connections suggest that learning and signal propagation are strongly influenced by the stability characteristics of the underlying neural PDEs, particularly in physics-informed settings.

Based on these considerations, we organize this paper as follows: 
We first examine the statistical properties of learned PINN weight matrices using viscous and inviscid variants of the one-dimensional Burgers’ equation, and compare them with predictions from random matrix theory. 
We then analyze how random and structured weight matrices influence the stability of signal propagation in deep networks. 
Finally, we interpret these observations within the neural differential equation framework, and demonstrate how numerical stability of the underlying neural PDEs governs the resulting network dynamics.  Through this analysis, we aim to provide a unified perspective linking weight statistics, signal propagation, and numerical stability in physics-informed neural networks.


\section{Inspecting the weights of PINN solutions}
\subsection{Weight distribution}
In this section, we inspect the weights of a Physics Informed Neural Network (PINN), trained separately to learn two related problems, namely the viscid and inviscid Burgers' equation, that are widely used by the fluid dynamics community, known to have discontinuous solutions. The details of the problem itself are given in Tab.~\ref{tab:pde_ic_bc_summary}. For the inviscid case, a Riemann-type initial condition is used, to make it harder for the network to learn the problem.

\begin{figure}[htb]
\centering
\includegraphics[scale=0.4]{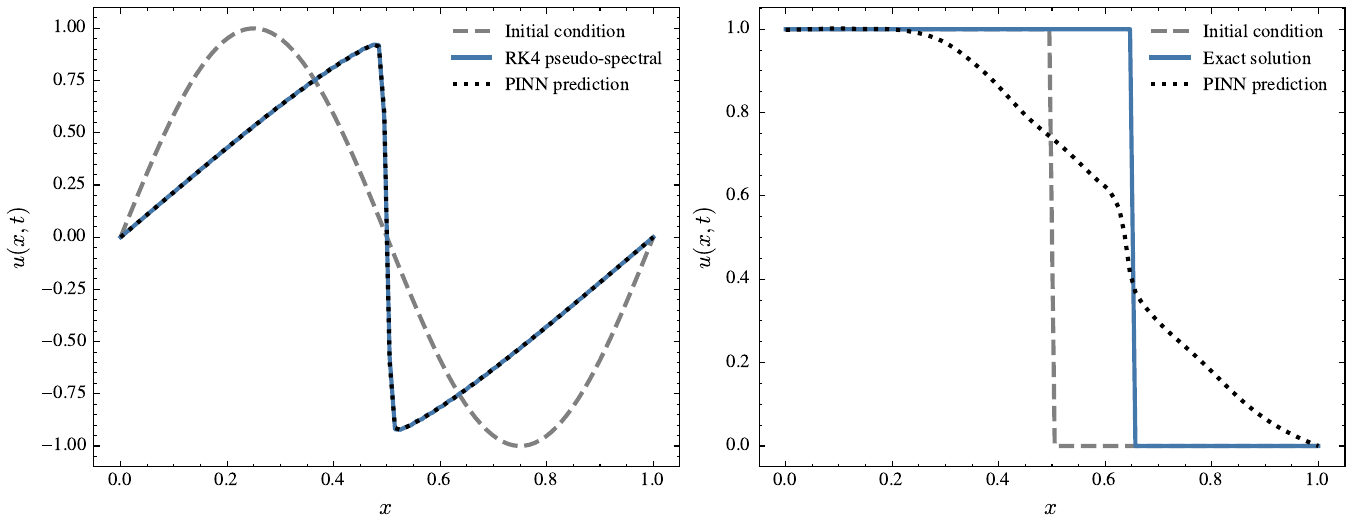}
\caption{The formation of a shock front under viscous and inviscid Burgers dynamics at $t = 0.3$. The left panel shows the PINN predicted solution for viscous variant, compared with the RK-4 pseudo-spectral method for reference. The right panel shows the inviscid case and the failure of PINN to learn the correct dynamics.}
\label{fig:nn_burgers_sol}.
\end{figure}

The ML procedure employs 8 hidden layers with 100 neurons each and $tanh$ as an activation function (except the output layer, which uses linear activation). 
For the viscous Burgers' equation, the viscosity $\nu$ is taken as $0.01/\pi$. Both the equations are solved for $x \in [0,1]$. The network is trained over 1 full convection time for both the problems. It can be seen from Fig.~\ref{fig:nn_burgers_sol} that the present network performs well in prediction of the shock for the viscous case, but fails to learn and give a satisfactory solution for the inviscid case. Convergence is not achieved for the inviscid case for even longer training times. Since the problem is exquisitely physical, one may expect
that the weights of an explainable ML scheme should reflect somehow the physical symmetries of the problem.

\begin{table}[ht]
\centering
\renewcommand{\arraystretch}{1.1}
\small
\caption{Initial boundary conditions for the numerical solution of the tested PDEs.}
\begin{tabular}{lccc}
\hline
\textbf{PDE} & \textbf{Conservative form} & \textbf{IC ($t=0$)} & \textbf{BC} \\
\hline
Viscous Burgers' &
$\dfrac{\partial u}{\partial t} + u\dfrac{\partial u}{\partial x} = \nu \dfrac{\partial^2 u}{\partial x^2}$
&
$u=sin(2\pi x)$
& $u(0,t)=0$ \\
& & & $u(1,t)= 0$ \\
\hline
Inviscid Burgers' &
$\dfrac{\partial u}{\partial t} + u\dfrac{\partial u}{\partial x} = 0$
&
$u = \begin{cases}
1 & x<0.5 \\
0 & x\ge0.5
\end{cases}$
& $u(0,t)=1$ \\
& & & $u(1,t)= 0$ \\
\hline
\end{tabular}
\label{tab:pde_ic_bc_summary}
\end{table}

\begin{figure}[hbtp]
\centering
\includegraphics[width=0.8\linewidth]{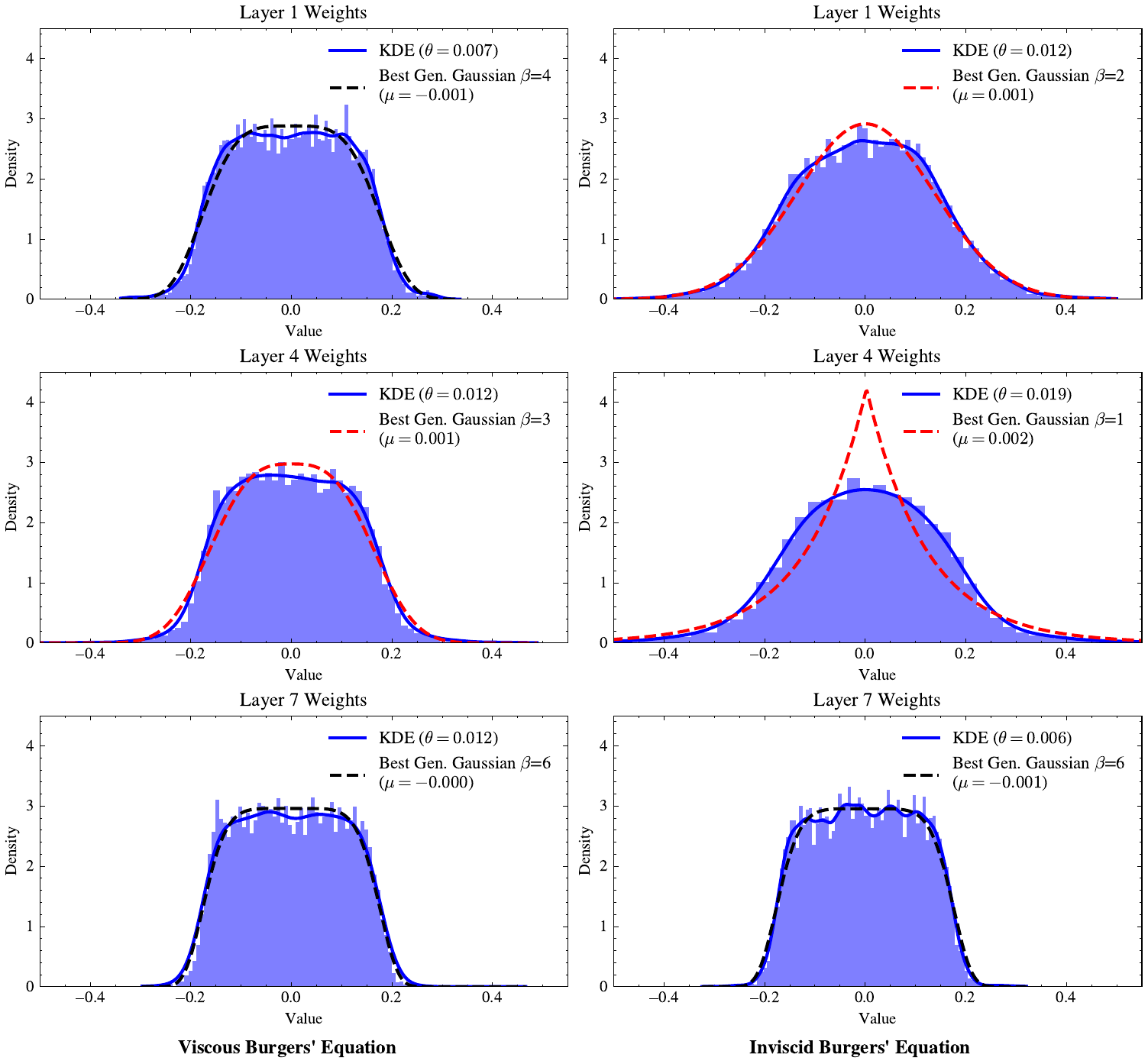}
\caption{The figure shows KDE-fitted probability density functions (PDFs) and MLE fits (via negative log-likelihood) for the PINN weight matrices trained on the Burgers equation over discrete, integer $\beta$ values as representative distributions. Panels correspond to weights from early, middle, and late hidden layers. The generalized Gaussian PDFs are parameterized by mean $\mu$ and standard deviation $\sigma$. Notably for the inviscid case, no generalized Gaussian provides a good fit (see Sec.~\ref{asec:fits}) for the early and middle layers; the MLE is dominated by the distribution tails, indicating a heavy-tailed structure consistent with implicit self-regularization~\cite{MartinMahoney2018}.
}
\label{fig:weights_biases}
\end{figure}

The weight distributions are shown in Fig.~\ref{fig:weights_biases}. We first obtain the Probability Distribution Function (PDF), or $p(x)$ by the means of Kernel Density Estimation, using cross validation to get 
optimal $\theta$ of the Gaussian fitting kernel $k(x)$ given by,
\begin{equation}
     k(x) \propto \exp\left(-\frac{(x-\hat\mu)^2}{2\theta}\right).
\end{equation}

We immediately see that the learned weights of the layers are fairly well-distributed around $0$ for all the layers considered for both the problems. Starting with a uniform distribution, the weights for both viscous and inviscid cases seem to settle at general Gaussian distributions, given by $\left(\text{i.e, } p(x) \propto\, e^{-\frac{|x-\mu|}{\alpha}^\beta}\right)$ with different, discrete shape parameter $(\beta)$ values determined by MLE (Maximum Likelihood Estimate). 

The weights of layer 4 in the inviscid case are particularly notable, with the MLE fit suggesting a super-Gaussian distribution $(\beta < 2, \text{e.g., Laplace with }\beta=1 )$, consistent with the kurtosis values reported in Tab.~\ref{tab:kurtosis}. However, it is visually clear from Fig.~\ref{fig:weights_biases} that such a fit is unsatisfactory and the tail-heavy behavior influences the MLE and kurtosis, pointing towards the existence of internal correlations. Albeit less obvious, similar can be seen for the layer 1 of inviscid case and layer 4 of viscid case as well. The full information about the fits can be found in Sec.~\ref{asec:fits}. Since the shape parameter $\beta$ is not intrinsically restricted to integer values, we also perform a continuous optimization over $\beta \in [1,20]$ in Sec.~\ref{asec:fits}, confirming that the discrete scan preserves the qualitative nature of the fits for the present discussion. 

\begin{table}[htbp]
\centering
\begin{tabular}{c c c c c}
\hline
 & \multicolumn{2}{c}{Viscid Burgers} & \multicolumn{2}{c}{Inviscid Burgers} \\
\hline
Layer & Weights Kurtosis & Biases Kurtosis & Weights Kurtosis & Biases Kurtosis \\
\hline
1 & 2.14 & 1.93 & 2.92 & 2.76 \\
4 & 2.94 & 5.8 & 6.47 & 3.01 \\
7 & 1.97 & 2.39 & 1.99 & 2.09 \\
\hline
\end{tabular}
\caption{Kurtosis of weights and biases across selected deep layers for viscid and inviscid Burgers equations (Normal/Gaussian baseline = 3).}
\label{tab:kurtosis}
\end{table}

\subsection{Spectral Analysis}
\begin{figure}[htb]
\centering
\begin{subfigure}{0.48\textwidth}
    \centering
    \includegraphics[scale=0.42]{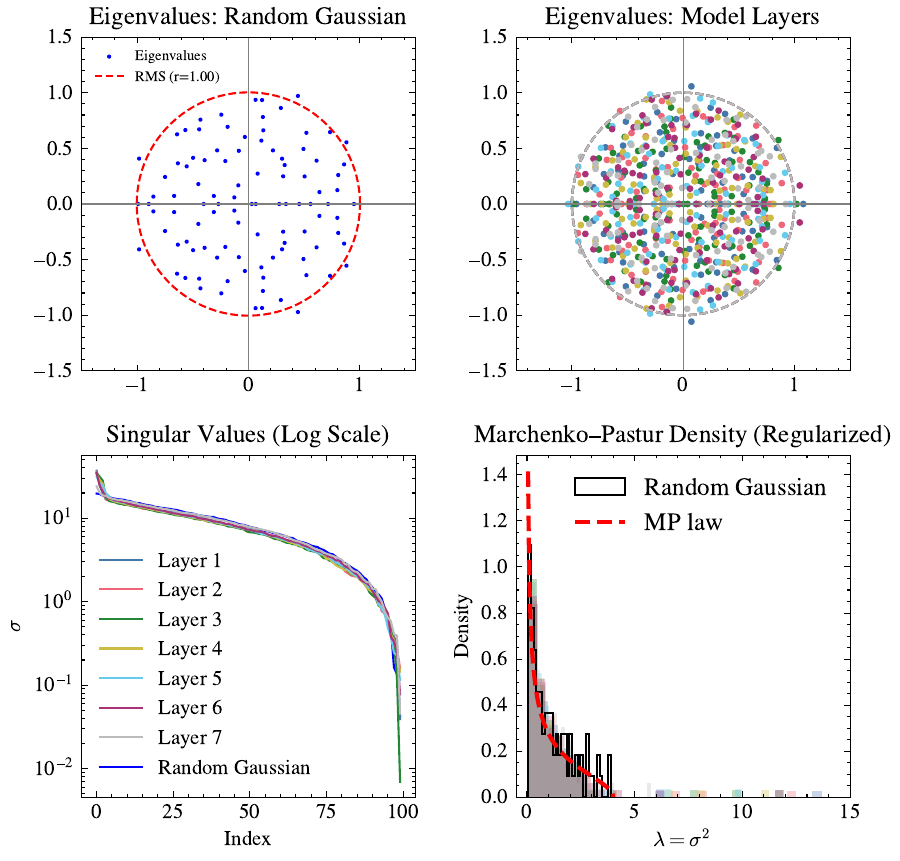}
    \caption{1D viscous Burgers' equation}
    \label{fig:1d_visc_eigenspec}
\end{subfigure}
\hfill
\begin{subfigure}{0.48\textwidth}
    \centering
    \includegraphics[scale=0.42]{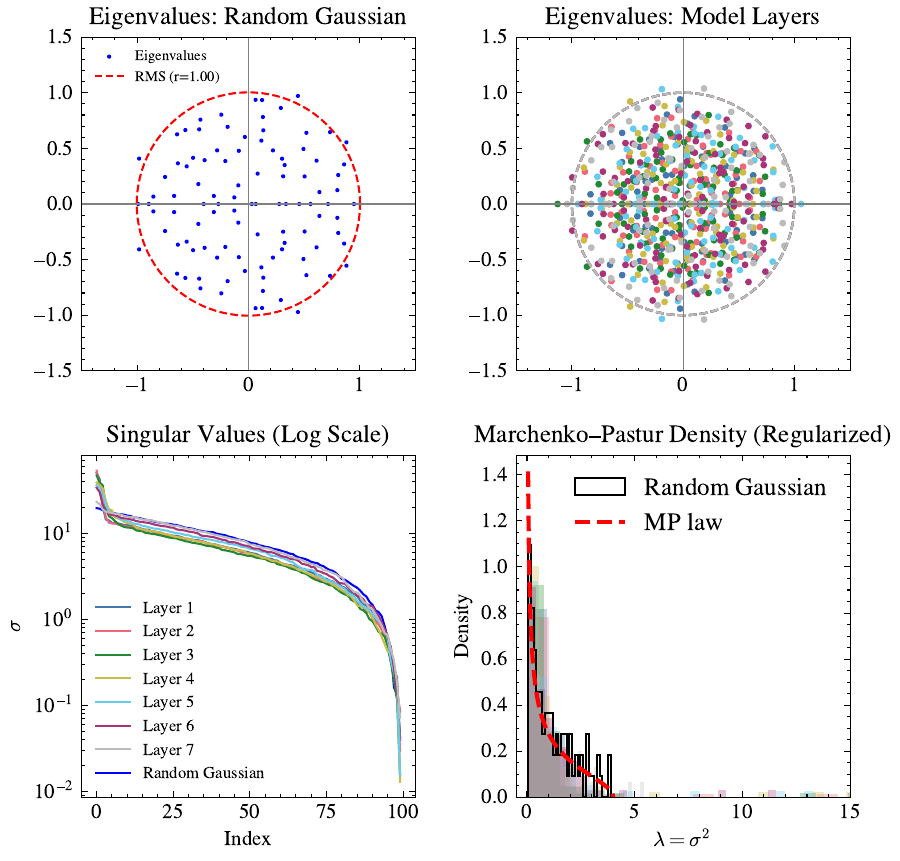}
    \caption{1D inviscid Burgers' equation}
    \label{fig:1d_invisc_eigenspec}
\end{subfigure}
\caption{
Spectral analysis of network weight matrices for (a) 1D viscous Burgers’ equation and (b) 1D inviscid Burgers’ equation.
Top row: Eigenvalue distributions of a random Gaussian matrix (left) and trained network layers (right), illustrating agreement with the circular law.
Bottom left: Singular values of the weight matrices across layers compared with a random Gaussian reference.
Bottom right: Empirical density of the squared singular values $(\sigma^2)$ of the normalized covariance matrix $WW^{T}$, compared with the Marchenko-Pastur (MP) law.}
\label{fig:eigen_values_burger}
\end{figure}

We then apply tools from random matrix theory (RMT) to examine the random-matrix-like properties of the weight matrices.
First, we plot the eigenvalues of the normalized weight matrices $W$ across 
all layers (see Fig.~\ref{fig:eigen_values_burger}) for the two equations.
We see both in Figs.~\ref{fig:1d_visc_eigenspec} and \ref{fig:1d_invisc_eigenspec} that the eigenvalues follow the 
circular theorem to an agreeable extent, suggesting independence and randomness in the weight matrices. For the inviscid  problem in Fig.~\ref{fig:1d_invisc_eigenspec}, the eigenvalues appear to be more clustered towards the center than the viscous counterpart.

We investigate this further in Fig.~\ref{fig:eigen_values_burger} by plotting the singular values of the normalized weight 
matrices and compare it with those of a random Gaussian matrix, which is done using SVD decomposition, $W = U\,\Sigma\,V^\dagger$.
Here, $U$ and $V$ are the left and right singular vectors, and $\Sigma$ represents the 
diagonal matrix, with the singular values $\sigma_i$  as its diagonal elements.
It can be observed that the qualitative trend significantly matches that of 
the random Gaussian matrix (matrix with elements drawn from a Gaussian distribution $\mathcal{N}(1,0)$), though with a few differences. 

First point of difference is the steep ``drop'' for both the problems near the highest singular 
values (leftmost point curve), a feature missing 
in the Gaussian counterpart. This may point to a slight directional preference by the post-training weight matrices as they try to adapt to a solution. Another caveat is that the overlay with the random Gaussian singular values is lesser for the inviscid case as compared to the viscous counterpart. 

Next, we compare the density distribution of the eigenvalues for the normalized covariance matrix $WW^\dagger$. While the majority trend of this feature for the trained weights are in well agreement with the Marchenko-Pastur law, there exist outliers in the weight matrix that fall outside this distribution. This observation may be attributed to the finiteness of the network size, combined with the fact that the learning of the problem itself introduces some hidden features in the weights. Here as well for the inviscid case, the higher eigenvalues tend to show visible deviation from the predicted MP density for random matrices.

It is interesting here to see that the differences in the statistics of the weight matrices from the RMT arise more significantly for the inviscid case, where the network has failed to learn the problem. For the viscid case, random matrix statistics are followed more agreeably by the learned weights. While a deeper analysis is surely required at this stage, there seems to be a preferred randomness-dominated nature of the well-trained weight matrices with seemingly little or no ``memory'' of the mathematical 
structure of the physical problem under study, thereby casting doubts
on the problem-specific explainability and insight. 
This said, the random-like nature of the PINN matrix invites a further observation. PINNs may represent a kind of (generalized) path integral 
solution of the PDE in point. This connection at present is speculative in nature and future work will explore precise technical details.

\section{Effect of random weights on signal propagation}
\begin{figure}[ht]
\centering
\begin{subfigure}{0.4\textwidth}
    \centering
    \raisebox{0.5\height}{%
        \includegraphics[scale=0.4]{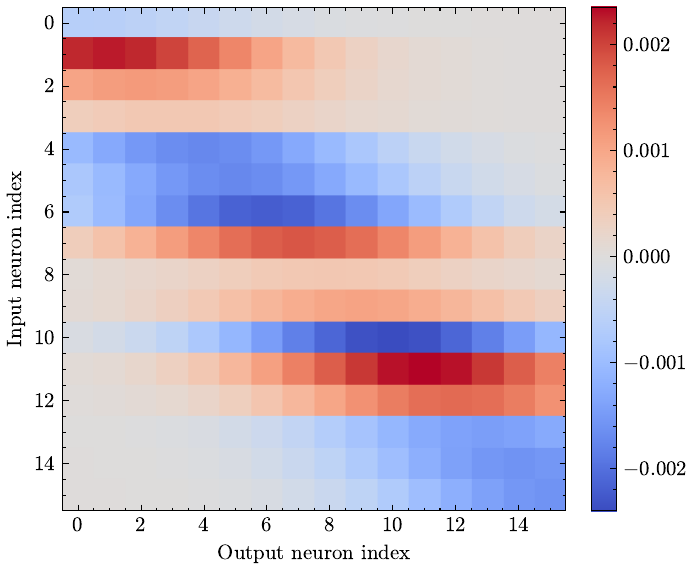}
    }
    \caption{}
    \label{fig:kernel_Gaussian_diagonal}
\end{subfigure}
\hfill
\begin{subfigure}{0.58\textwidth}
    \centering
    \includegraphics[scale=0.7]{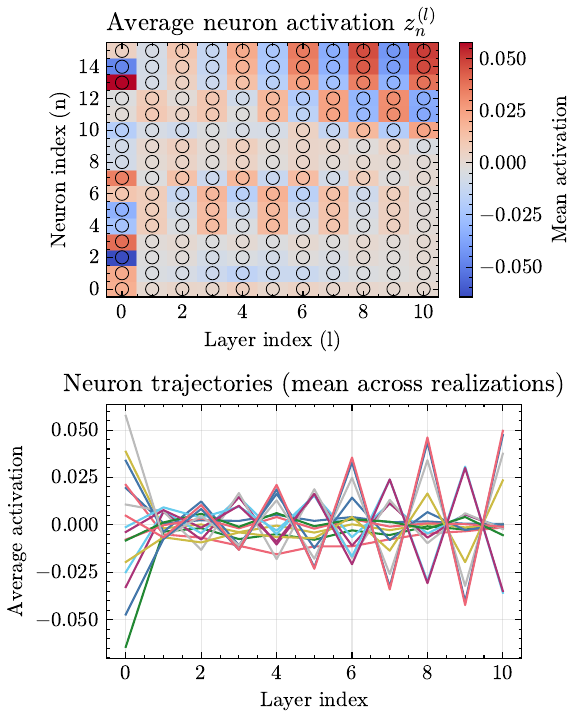}
    \caption{}
    \label{fig:neuralchain_Gaussian_diagonal}
\end{subfigure}
\caption{(a) A random Gaussian kernel with 
    diagonal scaling, corresponding to a diagonally dominant weight matrix; (b) The signal evolution over $N_R=5$ independent runs for the diagonally-scaled Random Gaussian kernel. 
    The top panel shows the heat maps of the average activation of each neuron, while the bottom 
    panel shows the trajectories of the signal features through the network. The model has 10 layers, with 16 neurons each.} 
\end{figure}

Having shown the inherent randomness in the trained weights, we demonstrate the effects of random weights on the signal propagation inside the network in this section. Given an input signal which transforms to have $n$ features (or an $n-$ dimensional vector, $\pmb{z}$) in the first hidden layer, the evolution over depth may be modeled as:
\begin{equation}
z_n^{(l)} = \sigma\!\left( \sum_{m} W_{nm}^{(l)} z_m^{(l-1)} + b_n^{(l)} \right),
\end{equation}
where  $W$ is a square weight matrix owing to equal number of neurons in each hidden layer. We consider a partially deterministic kernel in the form of a diagonally dominant row-wise random Gaussian kernel, similar to those employed in radial basis function (RBF) interpolation. The weight matrix is generated from a row-wise Gaussian kernel and normalized by its diagonal entries, thereby enforcing diagonal dominance. It can be visualized by Fig.~\ref{fig:kernel_Gaussian_diagonal}. Here, \textit{tanh} activation function is used in between the steps.

Fig.~\ref{fig:neuralchain_Gaussian_diagonal} shows the evolution of random signals, averaged over $5$ independent runs inside a hypothetical network of 
$L=10$ layers with $N=16$ neurons each, employing the weights from this kernel. 
The apparent instability in the evolution of the input signal is visible in form of oscillatory 
evolution as it passes through the network. It seems safe to say that the signal never converges even if the number of layers are increased. Hence, even a semi-deterministic kernel of weights doesn't guarantee learning of the network.

\section{Link to neural PDE and numerical stability}
We now examine the role of weight matrices in deep neural networks from the perspective of neural partial differential equations (neural PDEs). 
It has been shown in~\cite{NPDE} that weight kernels can be interpreted as discrete generators of PDE-like dynamics governing signal propagation across layers. 
Such neural PDEs have been demonstrated to accurately model a class of deep networks with uniform hidden-layer widths.
\begin{figure}[ht]
    \centering
    \includegraphics[width=0.3\linewidth]{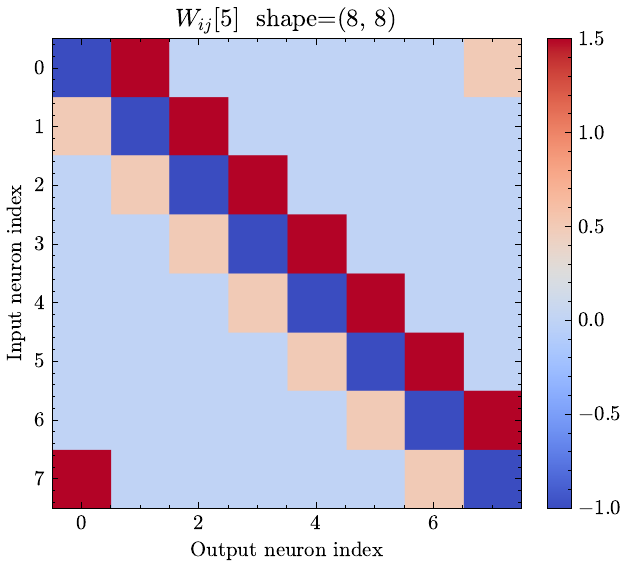}
    \caption{The representation of the weight matrix obtained from Eq.~\ref{eq:weights_adv_diff}, with imposed periodic boundary.}
    \label{fig:TDM_weight}
\end{figure}

We consider the advection-diffusion neural PDE introduced in~\cite{succi2026neuralchains}, whose discrete weight structure is given by
\begin{equation}
\label{eq:weights_adv_diff}
W_{ij} = (D+U/2)\,\delta_{i-1,j} + (1-2D)\,\delta_{i,j} + (D-U/2)\,\delta_{i+1,j},
\end{equation}
where $D$ denotes the diffusion coefficient and $U$ the advection velocity. 
To close the system, we impose periodic boundary conditions on $W_{ij}$, which provide a convenient and physically meaningful framework for analysis, despite the abstract mathematical origin of the neural PDE. The signal propagation through the network corresponds to the time integration of the associated neural PDE, particularly when linear activations are employed.

\begin{figure}[ht]
\centering
\includegraphics[width=0.8\linewidth]{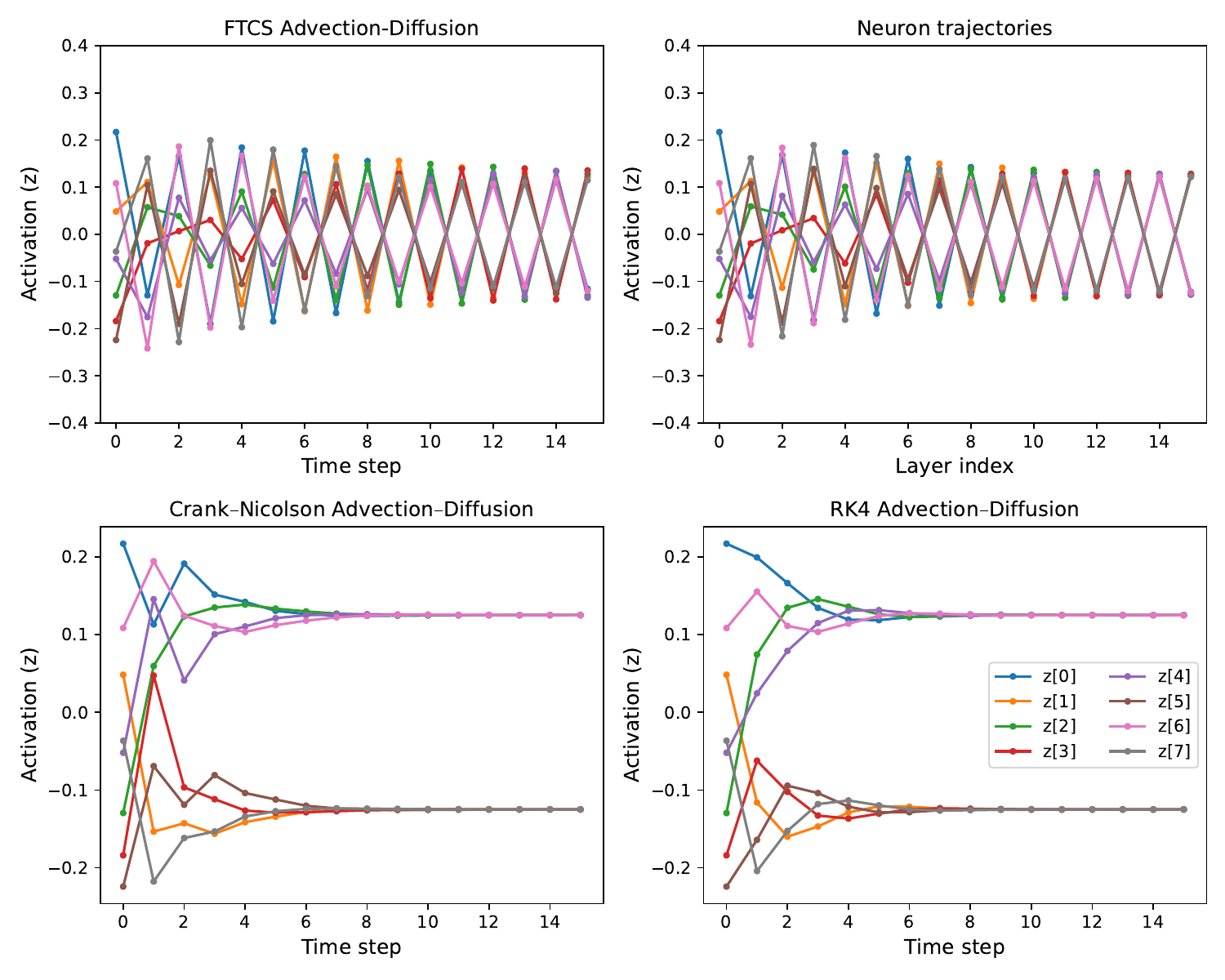}
\caption{Comparison between signal propagation in an 8-neuron network and the numerical solution of the corresponding advection–diffusion neural PDE, assuming trained network weights.
Top: Unstable evolution of a signal in the neural network (left) and a very similar numerical solution of the corresponding neural PDE, obtained using an explicit finite-difference scheme (right).
The bottom row demonstrates stable propagation when implicit time-integration methods are employed, specifically Crank–Nicolson (left) and fourth-order Runge–Kutta (RK4, right).
}

\label{fig:fdm_vs_neuralevol}
\end{figure}

Let us consider the case $U = D = 1$ and a network comprising 15 layers with 8 neurons per layer (see Fig.~\ref{fig:TDM_weight} for the weight matrix). We first solve the corresponding neural PDE using an explicit finite-difference scheme with $dt = dx = 1$ and periodic boundary conditions. 
As shown in the right panel of Fig.~\ref{fig:fdm_vs_neuralevol}, the resulting numerical solution exhibits classical instability in the form of checkerboard oscillations.

The left panel of Fig.~\ref{fig:fdm_vs_neuralevol} displays the evolution of an input signal in the corresponding neural architecture. 
Remarkably, a similar unstable behavior is observed here, even in the presence of nonlinear activation function ($tanh$ here). 
This correspondence indicates that the network inherits the numerical stability properties of the underlying discretized neural PDE.

In contrast, implicit and higher-order time-integration schemes, such as the Crank--Nicolson and Runge--Kutta methods, yield stable solutions for the same neural PDE, as illustrated in the bottom panel of Fig.~\ref{fig:fdm_vs_neuralevol}. 
This observation suggests that stability properties of numerical schemes play a fundamental role in determining the reliability of signal propagation in neural chains.
More generally, these results indicate that the behavior of deep networks is influenced not only by optimization and expressivity, but also by the stability properties of the underlying neural PDEs they implement. 
Similar considerations arise in deep equilibrium models and neural ODEs, where convergence governs network dynamics.

Accordingly, the neural PDE framework provides a useful perspective for interpreting signal propagation in deep networks, and suggests that stability properties are reflected in the structure of the learned weight matrices.

\section{Conclusion and outlook}

In this work, we presented a statistical and spectral analysis of trained PINN weights for viscous and inviscid variants of the one-dimensional Burgers’ equation and showed that the learned weights tend to approach (and prefer) a high-entropy regime consistent with predictions from random matrix theory. We then showed the evolution of signal features inside a network through the lens of neural partial differential equations (neural PDEs) to investigate the dynamical consequences of such weight structures.  We show that random and structured weight matrices can be associated with specific discretizations of neural PDEs, and that the numerical stability of these discretizations governs the stability of signal propagation through the network. 
We demonstrated that the same underlying neural PDE shows unstable or well-behaved dynamics when solved using explicit or implicit (or higher-order explicit) schemes, respectively. We note that the standard sequential PINN architecture effectively behaves like an explicit scheme, in which each layer depends only on the previous one. As we show, this explicit-like propagation can be numerically unstable for the underlying neural PDE and architecture under consideration.

\section*{Acknowledgements}
The author SS is grateful to the Simons Foundation for supporting several enriching visits of his. 
He also wishes to acknowledge many enlightening discussions 
over the years with PV Coveney, A. Laio, G. Longo and D. Spergel.
The authors also grateful to M. Durve for critical reading of the manuscript and
S. Strogatz for valuable comments. 
Finally they wish to acknowledge many insightful critical remarks by A. Gabbana. 

\section*{Funding}
SA acknowledges support from the Abdul Kalam Technology Innovation National Fellowship (INAE/SA/4784). JMT is grateful to the FRQNT "Fonds de recherche du Québec - Nature et technologies (FRQNT)" for financial support (Research Scholarship No. 357484).
SS  gratefully acknowledges funding by the European Union (EU) under the Horizon Europe research and innovation programme, EIC Pathfinder - grant No. 101187428 (iNSIGHT)

\section*{Conflict of interest}
The authors have no relevant financial or non-financial interests to disclose.

\section*{Data availability}
The scripts along with the result datasets are
available at \url{https://github.com/abhishekganguly808/randomness-in-PINNs}

\bibliography{eoi}
\newpage
\appendix
\section{Weight kernel drawn from uniform distribution}
\begin{figure}[ht]
    \centering
    \includegraphics[width=0.85\linewidth]{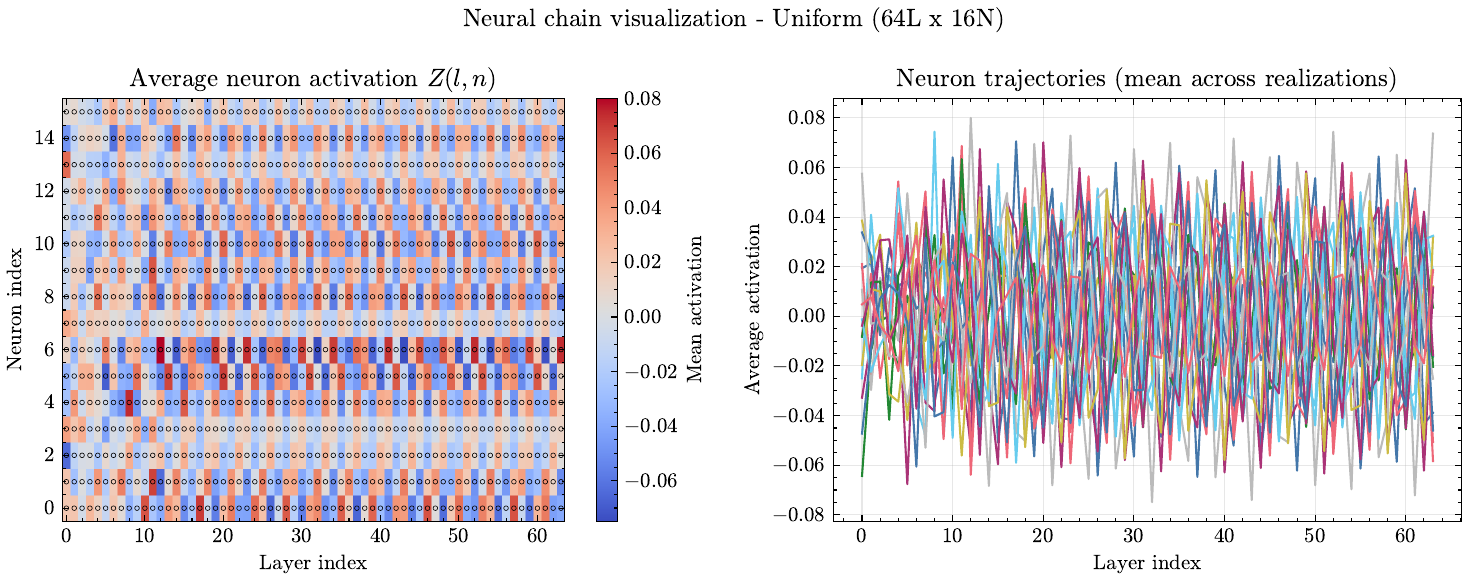}
    \vspace{1em} 
    \includegraphics[width=0.85\linewidth]{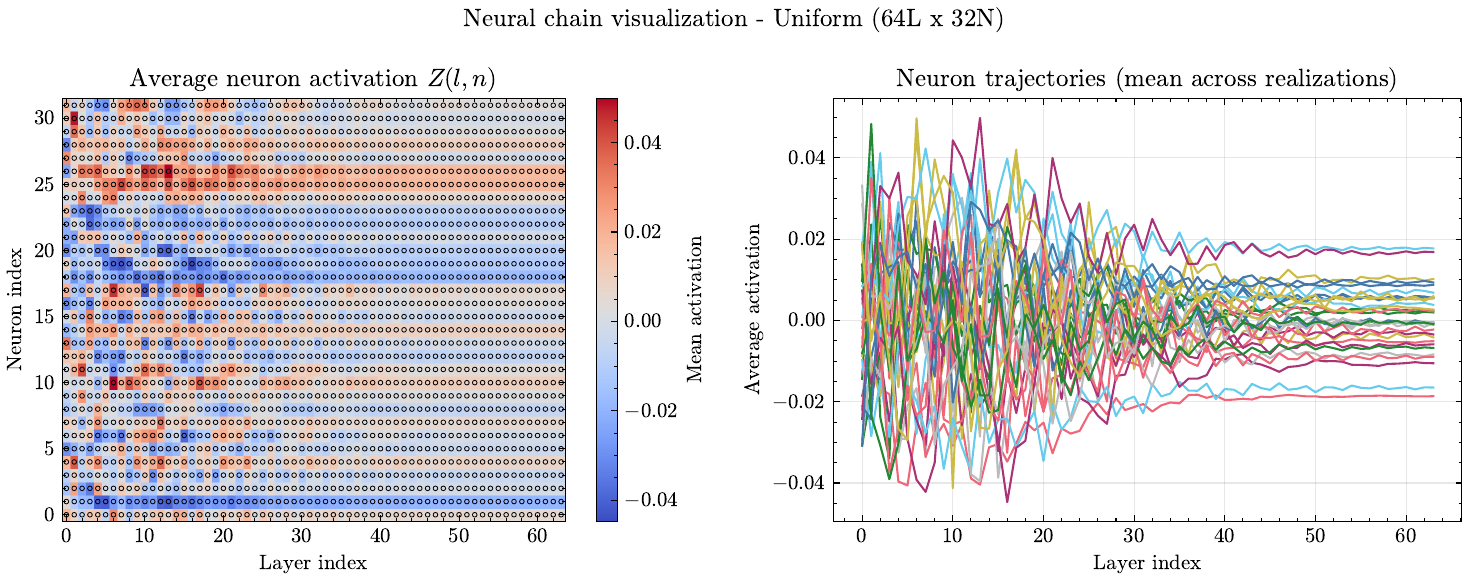}
    \caption{Neuron activation evolution for (top) 64 layer, 16 neuron per layer network and (bottom) 64 layer, 32 neurons per layer network. The top panels show an unstable evolution of a signal with 16 features, whereas the bottom panel shows a stable convergence of an initial state with 32 features.} 
    \label{fig:neuralchain_uniform}
\end{figure}

We show that the instability in training or failure to learn does not always find the weighst as the culprit. And show that the network architecture has a role to play as well. Let us now take a kernel of uniform distribution.
We take two networks, as shown in Fig.~\ref{fig:neuralchain_uniform}. 
It becomes apparent  that the evolution of the input signal is sensitive to the choice of
the number of neurons per layer/independent features of the signal. 
For the same $L=64$ layer network, $N=16$ neuron per layer case yields a checkerboard solution 
which is considered unstable, whereas with $N=32$ neurons per layer, the input signal converges fairly well.
This supports evidence of an underlying numerical system that is significantly sensitive to the  ``discretization'' of the network itself.

\section{Maximum Likelihood Estimation of Weight Distributions}\label{asec:fits}

The empirical distributions of the network weights are modeled using a generalized Gaussian distribution with mean $\mu$, standard deviation $\alpha$, and shape parameter $\beta$, whose probability density function reads
\begin{equation}
p(x;\mu,\alpha,\beta)
=
\frac{\beta}{2\alpha\Gamma(1/\beta)}
\exp\!\left[
-\left(\frac{|x-\mu|}{\alpha}\right)^{\beta}
\right].
\end{equation}

For a fixed value of $\beta$, the parameters $(\mu,\alpha)$ are estimated by maximizing the log-likelihood
\begin{equation}
\mathcal{L}(\mu,\alpha \mid \beta)
=
\sum_{i=1}^{N}
\log p(x_i;\mu,\alpha,\beta),
\end{equation}
where $\{x_i\}_{i=1}^N$ denotes the sampled weights. The optimization is performed using the L-BFGS-B algorithm, with $\alpha$ parameterized in logarithmic form to enforce positivity.

Since the shape parameter $\beta$ is not known \emph{a priori}, a discrete scanning procedure is employed over the range $\beta \in [1,8]$. For each candidate value, maximum likelihood estimates of $(\mu,\alpha)$ are computed and model selection is carried out using the Akaike Information Criterion~\cite{akaike1974aic},
\begin{equation}\label{aeq:aic}
\mathrm{AIC} = 2k - 2\mathcal{L}_{\max},
\end{equation}
with $k=2$. The optimal value of $\beta$ is determined by minimizing the AIC.

The selected generalized Gaussian distribution is finally compared with the kernel density estimate of the empirical distribution to assess the goodness of fit and to quantify deviations from Gaussian behavior. The data is tabulated in Tab.~\ref{tab:ggd_fit}.

\begin{sidewaystable}[t]
\centering
\small
\renewcommand{\arraystretch}{1.2}

\begin{tabular}{llcccccccc}
\toprule
Case & Metric
& $\beta=1$ & $\beta=2$ & $\beta=3$ & $\beta=4$
& $\beta=5$ & $\beta=6$ & $\beta=7$ & $\beta=8$ \\
\midrule

\multirow{3}{*}{Layer 1 (Viscous)}
& MLE
& 6603.78 & 7336.86 & 8020.59 & \textbf{8077.22}
& 8013.62 & 7883.45 & 7372.76 & 7524.91 \\

& AIC
& -13203.55 & -14669.71 & -16037.17 & \textbf{-16150.44}
& -16023.24 & -15762.91 & -14741.51 & -15045.82 \\

& $p$-value
& 0.00 & 0.00 & 0.00 & \textbf{0.09}
& 0.05 & 0.00 & 0.00 & 0.00 \\

\midrule

\multirow{3}{*}{Layer 1 (Inviscid)}
& MLE
& 5052.22 & \textbf{5696.51} & 5594.52 & 5207.24
& 4920.13 & 4552.02 & 4207.25 & -- \\

& AIC
& -10100.45 & \textbf{-11389.01} & -11185.04 & -10410.47
& -9836.27 & -9100.03 & -8410.49 & -- \\

& $p$-value
& 0.00 & \textbf{0.01} & 0.00 & 0.00
& 0.00 & 0.00 & 0.00 & 0.00 \\

\midrule

\multirow{3}{*}{Layer 4 (Viscous)}
& MLE
& 6434.80 & 7342.80 & \textbf{7387.67} & 7153.15
& 6854.50 & 6554.21 & 6297.77 & 6054.43 \\

& AIC
& -12865.61 & -14681.61 & \textbf{-14771.33} & -14302.30
& -13705.01 & -13104.43 & -12591.55 & -12104.85 \\

& $p$-value
& 0.00 & 0.00 & \textbf{0.01} & 0.00
& 0.00 & 0.00 & 0.00 & 0.00 \\

\midrule

\multirow{3}{*}{Layer 4 (Inviscid)}
& MLE
& \textbf{4413.98} & 4340.08 & 3895.82 & 2826.20
& 2528.88 & 1969.28 & 1521.18 & 1158.33 \\

& AIC
& \textbf{-8823.97} & -8676.15 & -7787.63 & -5648.40
& -5053.75 & -3934.56 & -3038.37 & -2312.65 \\

& $p$-value
& \textbf{0.00} & 0.00 & 0.00 & 0.00
& 0.00 & 0.00 & 0.00 & 0.00 \\

\midrule

\multirow{3}{*}{Layer 7 (Viscous)}
& MLE
& 7209.44 & 8433.26 & 8855.92 & 9035.72
& 9113.78 & \textbf{9135.63} & 9114.28 & 8628.81 \\

& AIC
& -14414.88 & -16862.53 & -17707.84 & -18067.43
& -18223.56 & \textbf{-18267.26} & -18224.57 & -17253.62 \\

& $p$-value
& 0.00 & 0.00 & 0.00 & 0.00
& 0.00 & \textbf{0.06} & 0.40 & 0.00 \\

\midrule

\multirow{3}{*}{Layer 7 (Inviscid)}
& MLE
& 7380.85 & 8492.71 & 8957.74 & 9113.70
& 9098.23 & \textbf{9117.38} & 9059.23 & 8984.27 \\

& AIC
& -14757.70 & -16981.41 & -17911.47 & -18223.41
& -18192.46 & \textbf{-18230.76} & -18114.46 & -17964.54 \\

& $p$-value
& 0.00 & 0.00 & 0.00 & 0.00
& 0.00 & \textbf{0.49} & 0.32 & 0.04 \\

\bottomrule
\end{tabular}
\caption{Maximum likelihood estimates (MLE), Akaike Information Criterion (AIC), and estimated Kolmogorov-Smirnov $p$-values for generalized Gaussian fits across network layers. Best models (minimum AIC) are highlighted in bold.}
\label{tab:ggd_fit}

\end{sidewaystable}

\begin{figure}[ht]
    \centering
    \includegraphics[width=0.85\linewidth]{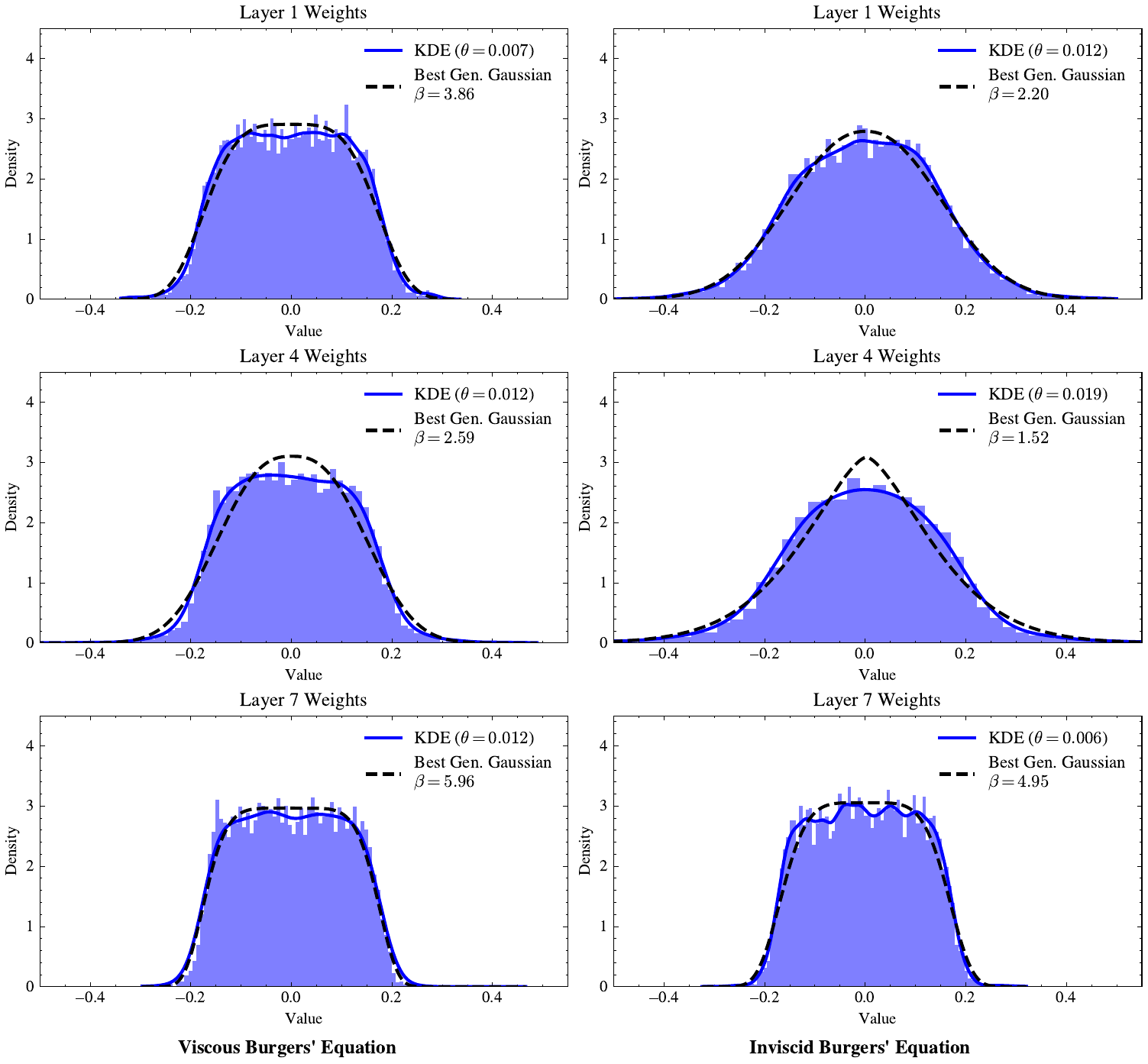}
    \caption{The figure presents kernel density estimates (KDEs) of the empirical weight distributions alongside MLE fits obtained by minimizing the negative log-likelihood. The weights correspond to early, intermediate, and late hidden layers of the PINN trained on the Burgers equation. The fitted distributions are generalized Gaussian probability density functions parameterized by the mean $\mu$, standard deviation $\sigma$, and shape parameter $\beta$. The parameter $\beta$ is optimized over the continuous range $\beta \in [1,20] \subset \mathbb{R}$.}
    \label{afig:weights_biases_cont}
\end{figure}

Furthermore, to assess the effect of restricting $\beta$ to integer values, we perform an additional fit in which the shape parameter $\beta$ is optimized continuously over the interval $[1,20]$. In this case, $\beta$ is treated as a free parameter, so the AIC penalty in Eq.~\ref{aeq:aic} uses $k=3$. The resulting MLE curves shown in Fig.~\ref{afig:weights_biases_cont} are not significantly different in terms of the statistical nature from those obtained by discrete scan shown in Fig.~\ref{fig:weights_biases}. The optimal continuous values of $\beta$ are reported in Tab.~\ref{tab:ggd_fit_cont}. 

\begin{table}[ht]
\centering
\begin{tabular}{lcccc}
\hline
\textbf{Layer} & $\boldsymbol{\beta}$ & \textbf{MLE} & \textbf{AIC} & $\boldsymbol{\beta}$(from discrete scan) \\
\hline
Layer 1 Viscous   & 3.87 & 8078.64 & -16151.29 & 4 \\
Layer 1 Inviscid  & 2.20 & 5706.28 & -11406.57 & 2\\
Layer 4 Viscous   & 2.59 & 7425.12 & -14844.24 & 3\\
Layer 4 Inviscid  & 1.52 & 4687.46 & -9368.93  & 1\\
Layer 7 Viscous   & 5.96 & 9135.87 & -18265.74 & 6\\
Layer 7 Inviscid  & 4.95 & 9151.68 & -18297.37 & 5\\
\hline
\end{tabular}
\caption{Maximum likelihood estimates and model selection metrics for selected layers under viscous and inviscid settings.}
\label{tab:ggd_fit_cont}
\end{table}

\end{document}